\documentclass[runningheads]{llncs}
\usepackage{graphicx}
\usepackage{xspace}
\usepackage{color}
\newcommand{\ie}{\textit{i}.\textit{e}.,\xspace}
\newcommand{\eg}{\textit{e}.\textit{g}.,\xspace}
\usepackage{enumerate}

\usepackage{hyperref}
\usepackage[frozencache,cachedir=.]{minted}
\usepackage{caption}
\usepackage{listings}
\newenvironment{code}{\captionsetup{type=listing}}{}

\usepackage{array}
\newcolumntype{C}[1]{>{\centering}m{#1}}

\usepackage{footnote}
\makesavenoteenv{tabular}
\makesavenoteenv{table}

\usepackage{float}
\usepackage{listings}

\newfloat{code}{htbp}{lop}
\floatname{code}{Listing}

\usepackage{subcaption}

\usepackage{amsmath,amssymb}
\usepackage{tikz-cd}
\newcommand{\customfootnotetext}[2]{{
  \renewcommand{\thefootnote}{#1}
  \footnotetext[0]{#2}}}

\begin{document}
%
\title{TERA: the Toxicological Effect and Risk Assessment Knowledge Graph}
%
%
\author{Erik B. Myklebust\inst{1,2} \and Ernesto Jim{\'{e}}nez{-}Ruiz\inst{2,3} \and Jiaoyan Chen\inst{4} \and \\Raoul Wolf\inst{1} \and Knut~Erik Tollefsen\inst{1,5}}

\authorrunning{E. B. Myklebust et. al.}
\institute{
Norwegian Institute for Water Research, Oslo, Norway \and
Department of Informatics, University of Oslo, Norway \and
City, University of London, United Kingdom \and
Department of Computer Science, University of Oxford, United Kingdom \and 
Faculty of Environmental Sciences and Natural Resource Management, Norwegian University of Life Sciences, \AA s, Norway
}

\maketitle              
\begin{abstract}\noindent
    Ecological risk assessment requires large amounts of chemical effect data from laboratory experiments. Due to experimental effort and animal welfare concerns it is desired to extrapolate data from existing sources.
    To cover the required chemical effect data several data sources need to be integrated to enable their interoperability. In this paper we introduce the  Toxicological Effect and Risk Assessment (TERA) knowledge graph, which aims at providing such integrated view, and the data preparation and steps followed to construct this knowledge graph. 
    We also present the applications of TERA for chemical effect prediction and the potential applications within the Semantic Web community. 
    %
    %
\end{abstract}



%
%

\setcounter{footnote}{0}
\section{Introduction}

Expanding the scope of ecological risk assessment models is a key goal in computational ecotoxicological research. However, the limiting factor in risk assessment is often the availability of toxicological effect data for a given compound and a given organism (species). 
The potential use of ten to hundreds of test organisms becomes ethically questionable. Moreover, collection of these data is
labour- and cost-intensive and often 
requires
extensive laboratory experiments.

One major challenge in risk assessment processes is the interoperability of data. 
In this paper we present the Toxicological Effect and Risk Assessment (TERA) Knowledge Graph that aims at providing an integrated view of the relevant data sources.\textsuperscript{$\dagger$}
\customfootnotetext{$\dagger$}{This paper focuses and extends on the construction of the TERA knowledge graph as a \textit{resource} for the ecotoxicological and Semantic Web domains, while our paper in \cite{Myklebust2019KnowledgeGE} had a special focus on the use of knowledge graph embeddings and machine learning for chemical effect prediction.}
%
The data sources that TERA integrates vary
from tabular, to RDF files and SPARQL queries over public linked data. 
Certain sources are very large and frequently updated, therefore TERA is materialized upon request via a series of APIs that are created to interact with TERA.

The main contributions of this paper are summarized as follows:
\begin{enumerate}[\it (i)]
    \itemsep0em 
    \vspace{-0.2cm}
    \item We have released the TERA knowledge graph. 
    A partially materialized snapshot among other relevant resources are publicly available (see Zenodo repository \cite{tera_kg}).
    \item TERA also includes the mappings between ECOTOX and NCBI for which there was not a complete and public alignment. 
    \item We have created a series of APIs to access, update and extend TERA (see GitLab repository \cite{tera_apis}).
    \item We describe several applications of the TERA knowledge graph both in the ecotoxicology domain (\eg chemical effect prediction) and the Semantic Web domain (\eg  embedding of hierarchical biased knowledge graphs, large scale ontology alignment).
\end{enumerate}

This paper is organized as follows. 
Section \ref{sec:ecotoxicology} provides an introduction to ecotoxicological risk assessment, while Section \ref{sec:preliminaries} provides some background to facilitate the understanding of the subsequent sections.
In Section \ref{sec:tera} we first present the data sources used to construct TERA, before we show the integration of these sources, and finally show several ways of accessing TERA. 
Section \ref{sec:applications} describes the potential applications of TERA, while Section \ref{sec:discussion} elaborates on the contribution of this work.

\section{Background}
\label{sec:ecotoxicology}

Ecotoxicology is a multidisciplinary field that studies the potentially adverse toxicological effects of chemicals on individuals, sub-populations, communities and ecosystems. In this context, risk is the result of the intrinsic hazards of a substance on species, populations or ecosystems, combined with an estimate of the environmental exposure, \ie the product of exposure and effect (hazard).

\begin{figure}[t]
    \centering
    \includegraphics[width=0.8\textwidth]{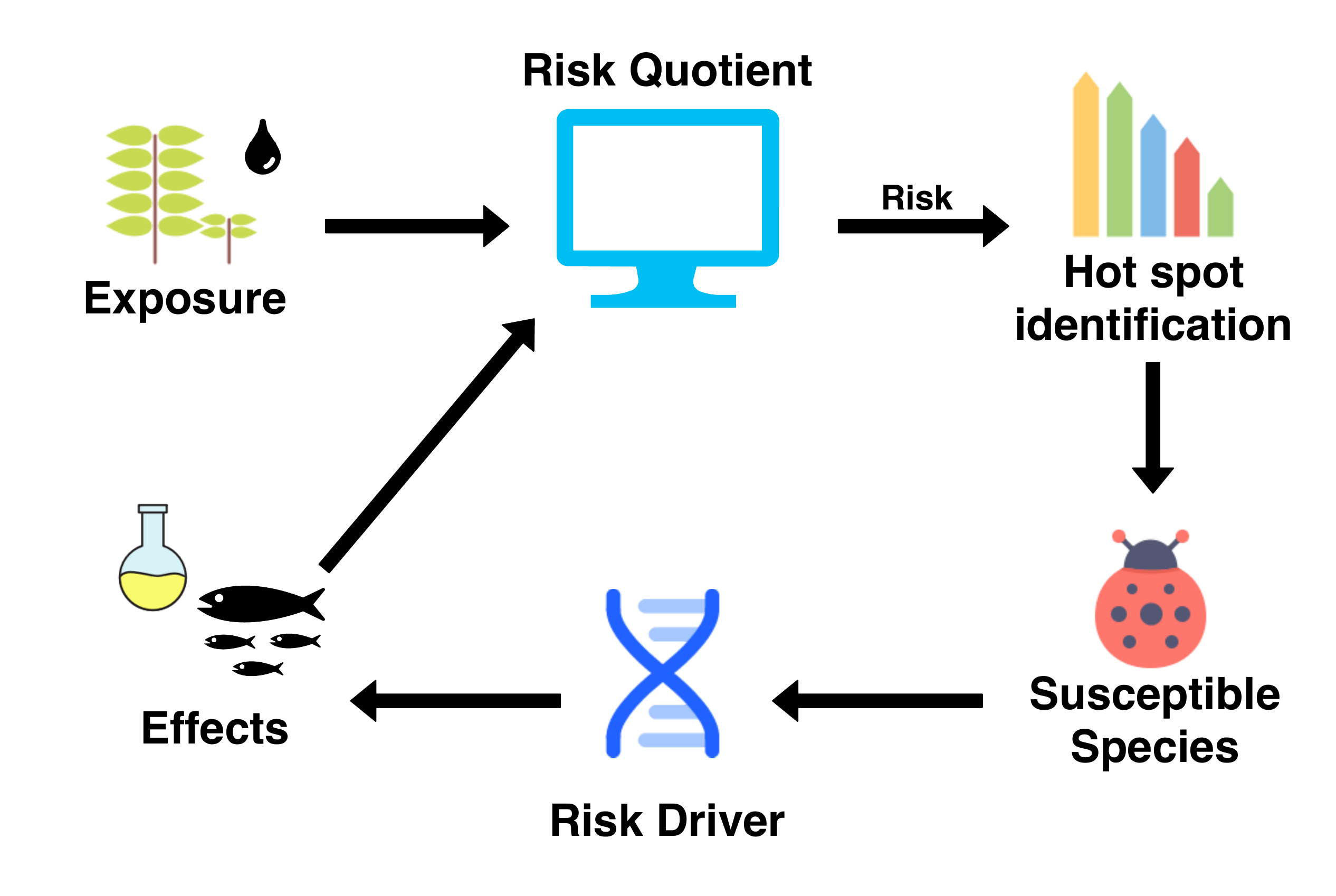}
    \caption{Simplified ecological risk assessment pipeline.}
    \label{fig:niva-pipeline}
\end{figure}

Figure \ref{fig:niva-pipeline} shows a simplified risk assessment pipeline.  %
\textit{Exposure} data is gathered from analysis of environmental concentrations of one or more chemicals, while \textit{effects} (\textit{hazards}) are characterized for a number of species in the laboratory as a proxy for more ecologically relevant organisms. These two data sources are used to calculate risk, using so-called assessment factors 
to extrapolate a risk quotient (RQ; ratio between exposure and effects). The RQ for one chemical or the mixture of many chemicals is used to identify chemicals with the highest RQs (risk drivers), susceptible species (or taxa), identify relevant modes of action\footnote{The functional or anatomical change in an organism due to exposure to a compound is called MoA.} (MoA) and characterize detailed toxicity mechanisms for one or more species (or taxa). Results from these predictions can generate a number of new hypotheses that can be investigated in the laboratory or studied in the environment. 

The effect data is obtained from available data, or in the case of no available data, during laboratory experiments, where the sub-population of a single species is exposed to a gradient of concentrations of a chemical. Most commonly, mortality rate, growth, development or reproductive output are measured over time.

To give a good indication of the toxicity to a species, these experiments are conducted with a concentration range spanning from no effect ($0\%$) to complete effect ($100\%$) when this is pragmatically possible. Hence some compounds will be more toxic 
than others and variance in susceptibility between species may provide a distribution of the effective concentration for one specific compound.

Ecological risk assessment requires large amounts of effect data to efficiently
predict
risk for the ecosystems and ecosystem components (\eg species and taxa). The data must cover a minimum number of the chemicals found when analysing environmental
samples, along with covering species and taxa present in the ecosystem. This leads to an immense search space that is close to impossible to encompass in its entirety and risk assessment is thus often limited by lack of sufficiently high quality effect data.
It becomes essential to extrapolate from known to unknown combinations of chemical--species pairs, which in some degree can be overcome by predicting the effects themselves through the use of
quantitative structure--activity relationship models (QSARs). 
These models have shown promising results for use in risk assessment, \eg \cite{Pradeep2016}, but have limited application domain (coverage), both in terms of compounds and species. 
Use of read-across and selection of proxy compounds that are chemically similar, display similar toxicity or have similar MoA and toxicity mechanisms are therefore becoming an 
attractive solution with increasing popularity (\eg \cite{doi:10.1002/qsar.200710099,WU201067}). Development of computational approaches that identify data that can be used for identifying proxy compounds to be used for read-across and data gap filling, is key to facilitate rapid, cost-effective, reliable and transparent predictions of new effects. 
We contribute in this regard by creating a semantic layer, \ie a knowledge graph, to enable extraction and integration of this high quality data.

\section{Preliminaries} 
\label{sec:preliminaries}

\medskip
\noindent
\textbf{Knowledge graphs}. 
We follow \cite{j.websem510} in the notion of a RDF-based knowledge graph which is represented as a set of RDF triples $\left\langle s, p, o \right\rangle$, where $s$ represents a subject (a class or an instance), 
$p$ represents a predicate (a property) 
and $o$ represents an object
(a class, an instance or a data value \eg text, date and number).
RDF entities (\ie classes, properties and instances) are represented by URIs (Uniform Resource Identifier). A knowledge graph consits of a terminology and an assertions box (TBox and ABox). The TBox is composed by RDF Schema constructs like class subsumption (\eg \texttt{ncbi:taxon/6668} \texttt{rdfs:subClassOf} \texttt{ncbi:taxon/6657}) and property domains and ranges (\eg \texttt{et:concentration} \texttt{rdfs:domain} \texttt{et:Chemical}).\footnote{The OWL 2 ontology language provides more expressive constructors. Note that the graph projection of an OWL 2 ontology can be seen as a knowledge graph (\eg \cite{jbs2018}).} The ABox contains relationships among instances and  type definitions (\eg \texttt{et:taxon/28868} \ \texttt{rdf:type} \ \texttt{et:Taxon}).

\medskip
\noindent
\textbf{SPARQL Queries. }
RDF-based knowledge graphs
can be accessed by SPARQL query language.\footnote{\url{https://www.w3.org/TR/rdf-sparql-query/}}
%
Next we summarise the SPARQL constructs used in this work:
\begin{enumerate}[\it (i)]
    \itemsep0em 
    \vspace{-0.1cm}
    \item Select queries are used when the desired output is tabular. 
    \item Construct queries can be used if the purpose of the query is to create a new graph. We use this in Listing \ref{lst:cas2inchikey} to create equivalence triples.
    \item \emph{Property paths} express multiple edges in a graph. \eg alternate paths (\eg \texttt{rdfs:label | foaf:name}), inverse relations (\eg \texttt{$\hat{~}$rdf:type}), path sequences (\eg \texttt{rdf:type /} \texttt{rdfs:subClassOf}), and any combination~of~these.
    \item A \emph{blank node} is a node where the identifier is not explicitly given. This allows the use of temporary nodes in queries. \eg Listing \ref{lst:example} uses \texttt{[\texttt{rdfs:label "Oslofjorden"@no}]} 
    to represent a node with label \emph{Oslofjorden}.
\end{enumerate}
Moreover, the extended syntax of SPARQL enables the use of complex property paths (\eg  a path of minimum 1 to maximum $n$ \texttt{rdfs:subClassOf} relations is represented as \texttt{rdfs:subClassOf\{1,n\}}), concatenating variables (\eg Listing \ref{lst:cas2inchikey}), aggregations and more.\footnote{\url{https://www.w3.org/wiki/SPARQL/Extensions}}

\medskip
\noindent
\textbf{Ontology alignment}. 
Finding the corresponding mappings between a source and a target ontology or knowledge graph is called ontology alignment~\cite{om2013}. In this work, computed mappings are represented in the knowledge graph as triples among the entities of the source and target (\eg \texttt{ncbi:taxon/13402} \texttt{owl:sameAs} \texttt{et:taxon/Carya}).

\section{The TERA Knowledge Graph}
\label{sec:tera}

This sections presents the data sources currently integrated within TERA, the APIs to prepare an integrate these data sources and the available entry points to access TERA.

\subsection{Data sources}
\label{sec:datasources}
The TERA knowledge graph is constructed from a number of sources,
including tabular data, RDF triples and SPARQL endpoints.


\medskip
\noindent
\textbf{Effect data.} 
The largest publicly available repository of effect data is the ECOTOXicology knowledgebase (ECOTOX) developed by the US Environmental Protection Agency \cite{ecotox}. This data is gathered from published toxicological papers and limited internal experiments.
The dataset consists of $940k$ experiments using $12k$ compounds and $13k$ species, implying a compound--species pair converge of maximum $\sim 0.6\%$. The resulting endpoint\footnote{Not to be confused with a SPARQL endpoint.} from an experiment is categorised in one of a plethora of predefined endpoints.
\eg for endpoints such as $EC50$ (\emph{effective concentration} on $50\%$ of test population), an
effect must be defined in conjunction with the endpoint. Mortality, chronic, and reproductive toxicity are common effect outcomes to characterise the \emph{effective concentration} of a compound upon a given target species.

\begin{table}[t]
    \setlength{\tabcolsep}{0.3pt}
    \centering
    \begin{tabular}{|c|c|c|c|}
        \hline
        test\_id & reference\_number & test\_cas & species\_number \\ \hline 
        1068553 & 5390 & 877-43-0 (2,6-Dimethylquinoline) & 5156 (\emph{Danio rerio}) \\
        2037887 & 848 & 79-06-1 (2-Propenamide) & 14 (\emph{Rasbora heteromorpha}) \\
        \hline
    \end{tabular}
    \medskip
    \caption{ECOTOX database tests examples.}
    \label{tab:ecotox_ex1}
\end{table}

\begin{table}[t]
    \centering
    \begin{tabular}{|c|c|c|c|c|}
        \hline
        result\_id & test\_id & endpoint & conc1\_mean & conc1\_unit \\ \hline 
        $98004$ & $1068553$ & $LC50$ & $400$ & $mg/kg$ diet \\
        $2063723$ & $2037887$ & $LC10$ & $220$ & $mg/L$ \\
        \hline
    \end{tabular}
    \medskip
    \caption{ECOTOX database results examples.}
    \label{tab:ecotox_ex2}
\end{table}

Tables \ref{tab:ecotox_ex1} and \ref{tab:ecotox_ex2} contains an excerpt of the ECOTOX database. 
ECOTOX includes information about the compounds and species used in the tests. 
This information, however, is limited and additional (external) resources are required to complement ECOTOX.  



\medskip
\noindent
\textbf{Compounds.}
The ECOTOX database use an identifier called CAS Registry Number assigned by the Chemical Abstracts Service to identify compounds. The CAS numbers are proprietary, however, Wikidata \cite{wikidata2014} (indirectly) encodes mappings between CAS numbers and open identifiers like \textit{InChIKey}, a 27-character hash of the International Chemical Identifier (InChI) which encodes chemical information uniquely\footnote{While InChI is unique, InChiKey is not, although collisions are few \cite{Willighagen2011}} \cite{Heller2015}.
Moreover, chemical features can be gathered from the chemical information dataset PubChem \cite{pubchem} using the open identifiers.
The classification of compounds in PubChem only concerns permutations of compounds. Therefore, we use the (Ch)EBI SPARQL endpoint to access the ChEMBL dataset, which enables us to create a more extensive classification hierarchy. 
To gather the functional properties of a chemical (\eg painkiller) we use the MeSH dataset, which is available from the MeSH SPARQL endpoint.

\medskip
\noindent
\textbf{Taxonomy.}
ECOTOX contains a taxonomy, however, this only considers the species represented in the ECOTOX effect data. Hence, to enable extrapolation of effects across a larger taxonomic domain, we introduce the NCBI taxonomy~\cite{ncbi}. This taxonomy data source
consists
of a number of database dump files, which contains a hierarchy for all sequenced species, which equates to around $10\%$ of the currently known life on Earth. For each of the taxa (species and classes), the taxonomy defines a handful of labels, most commonly used are the \emph{scientific} and \emph{common} names. However, labels such as \emph{authority} can be used to see the citation where the species was first mentioned, while \emph{synonym} is a alternate \emph{scientific} name, that may be used in the literature. 

\medskip
\noindent
\textbf{Species traits.}
As an analog to chemical features, we use species traits to expand the usability of the knowledge graph. The traits we have included in the knowledge graph are the habitat, endemic regions, and presence. This data is gathered from the Encyclopedia of Life (EOL) \cite{eol}, which is available as tabular files. Moreover, EOL uses external definitions of certain concepts, and mappings to these sources are available as glossary files.
In addition to traits, researchers may be interested in species that have different conservation statuses, \eg if the population is stable or declining, etc. This data can also be extracted from EOL.


\begin{figure}[t]
     \centering
     \includegraphics[width=0.7\textwidth]{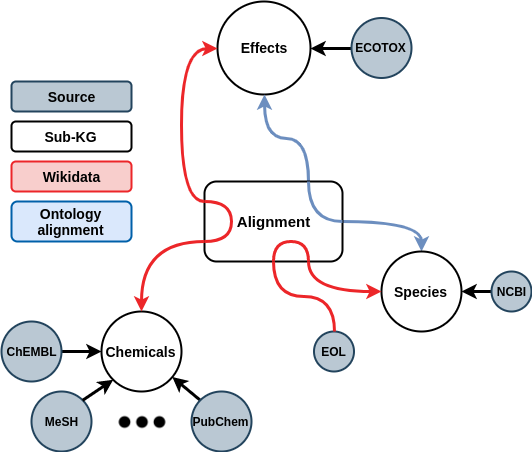}
     \caption{Data sources and colour-coded elements of the TERA knowledge graph.}
     \label{fig:datasources}
\end{figure}

\subsection{Preparing and Integrating Data into TERA}
\label{sec:datapreparation}


%
%

We have created four APIs for wrangling and incorporating effect, taxonomy, and chemical data into the TERA knowledge graph. These APIs also provide (predefined) methods to access the knowledge in TERA.
Figure \ref{fig:datasources} shows how the data sources integrate into the APIs and how the APIs map among each other.
Excluding the SPARQL endpoints,\footnote{Wikidata: \url{https://query.wikidata.org/sparql}\\ ChEMBL: \url{https://www.ebi.ac.uk/rdf/services/sparql}\\
MeSH: \url{https://id.nlm.nih.gov/mesh/query}
}
the data can be downloaded from the sources websites.\footnote{ECOTOX: \url{https://cfpub.epa.gov/ecotox/}\\ PubChem: \url{https://pubchemdocs.ncbi.nlm.nih.gov/downloads}\\ NCBI Taxonomy: \url{https://www.ncbi.nlm.nih.gov/guide/taxonomy/}\\
EOL: \url{https://opendata.eol.org/}
}
\par

\medskip
\noindent
\textbf{Species API.}
This API uses data from various tabular sources to describe the species taxonomy and related features. We use the namespace \url{https://www.ncbi.nlm.nih.gov/taxonomy} (\texttt{ncbi}) for the NCBI taxonomy.  

\begin{enumerate}
    \item {The integration of the the NCBI Taxonomy into the knowledge graph is split into several sub-tasks.
    \begin{enumerate}
        \item Loading the hierarchical structure included in \textit{nodes.dmp}. The columns of interest are the taxon identifiers of the child and parent taxon, along with the rank of the child taxon and the division where the taxon belongs. We use this to create triples like \texttt{(v)} and \texttt{(vi)} in Table \ref{tab:triples}.
        \item To aid alignment between NCBI and ECOTOX identifiers, we add the synonyms found in \textit{names.dmp}. Here, the taxon identifier, its name and name type are used to create triples similar to \texttt{(vii)} in Table \ref{tab:triples}. Note that a taxon in NCBI can have a plethora of synonyms while a taxon in ECOTOX usually have two, \ie common name and Latin name.
        \item Finally, we add the labels of the divisions found in \textit{divisions.dmp}. In addition, we add disjointness axioms among all divisions, \eg Triple \texttt{(ii)} in Table \ref{tab:triples}.
    \end{enumerate}
    }
    \item {
        The EOL traits data is available as tabular data, however using URIs, such that integrating the data is trivial. 
    }
\end{enumerate}

\begin{table}[t]
    \centering
    \begin{tabular}{|c c c c|}
        \hline
        \texttt{\#} & \texttt{subject} & \texttt{predicate} & \texttt{object} \\ \hline
        \texttt{(i)} & \texttt{et:group/Worms} & \texttt{owl:disjointWith} & \texttt{et:group/Fish} \\
        \texttt{(ii)} & \texttt{ncbi:division/2} & \texttt{owl:disjointWith} & \texttt{ncbi:division/4} \\
        \texttt{(iii)} & \texttt{ncbi:division/2} & \texttt{rdfs:label} & \texttt{``Mammals''} \\
        \hline 
        \texttt{(iv)} & \texttt{et:taxon/34010} & \texttt{rdfs:subClassOf} & \texttt{et:taxon/hirta} \\
        \texttt{(v)} & \texttt{ncbi:taxon/687295} & \texttt{rdfs:subClassOf} & \texttt{ncbi:taxon/513583} \\
        \texttt{(vi)} & \texttt{ncbi:taxon/687295} & \texttt{ncbi:rank} & \texttt{ncbi:Species} \\
        \texttt{(vii)} & \texttt{ncbi:taxon/687295} & \texttt{ncbi:scientific$\_$name} & \texttt{``Coleophora cornella''} \\
        \texttt{(viii)} & \texttt{ncbi:taxon/35525} & \texttt{eol:habitat} & \texttt{ENVO:00000873} \\
        \texttt{(ix)} & \texttt{ncbi:taxon/35525} & \texttt{eol:presentIn} & \texttt{worms:Oostende}
        \\\hline
        \texttt{(x)} & \texttt{et:test/001} & \texttt{et:compound} & \texttt{et:chemical/115866} \\
        \texttt{(xi)} & \texttt{et:test/001} & \texttt{et:species} & \texttt{et:taxon/26812} \\
        \texttt{(xi)} & \texttt{et:test/001} & \texttt{et:organsimLifestage} & \texttt{et:lifestage/adult}\\\hline
        \texttt{(xiii)} & \texttt{et:taxon/33155} & \texttt{owl:sameAs} & \texttt{ncbi:taxon/311871} \\
        \texttt{(xiv)} & \texttt{ncbi:taxon/311871} & \texttt{owl:sameAs} & \texttt{wd:Q13828695} \\
        \texttt{(xv)} & \texttt{et:chemical/115866} & \texttt{owl:sameAs} & \texttt{wd:Q418573} \\
        \hline
    \end{tabular}
    \medskip
    \caption{Example triples from the TERA knowledge graph.}
    \label{tab:triples}
\end{table}

\medskip
\noindent
\textbf{Chemical API.}
This API can be used either with local files downloaded from their respective sources (PubChem, ChEMBL, MeSH), or a local or online endpoints, depending on requirements. 
These chemical data sources are available as RDF and therefore integrating them into TERA is straightforward.

\begin{figure}[t]
    \centering
    \includegraphics[width=1.1\textwidth]{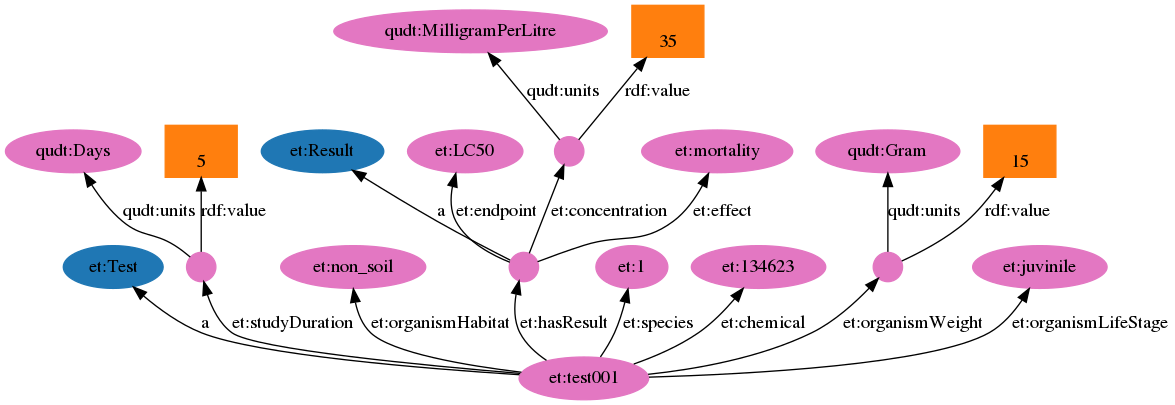}
    \caption{Example of a ECOTOX test and related triples.}
    \label{fig:ecotox_kg_ex}
\end{figure}

\medskip
\noindent
\textbf{Effect API}
The tabular data in ECOTOX requires significantly more cleaning than the other data. We use the namespace \url{https://cfpub.epa.gov/ecotox/} (\texttt{et}) for this part of TERA. 
\begin{enumerate}
    \item ECOTOX contains metadata about the species and compounds used in the experiments. We use this information to aim alignment between the effect and the background data. 
    \begin{enumerate}
        \item Species metadata in \emph{species.txt} include common and Latin names, along with a (species) ECOTOX group. This group is a categorization of the species based on ECOTOX use cases. We filter the species names, \eg \emph{sp.}, \emph{var.} (\ie unidentified species and variant) are removed along with various missing value shorthands used in the metadata.
        \item The full hierarchical lineage is also available in the \emph{species.txt} file.
        Each column represent a taxonomic level, \eg \emph{genus} or \emph{family}. If a column is empty, we construct a intermediate classification, \eg say \emph{Daphnia magna} has no genus classification in the data, then its classification will be Daphniidae genus (family name + genus, actually called \emph{Daphnia}). We construct these classifications to ensure the number of levels in the taxonomy is consistent. This consistency will help when aligning to the NCBI data. Note that when adding triples such as \texttt{(iv)} in Table \ref{tab:triples}, we also add a classification based on the column to aid easier querying for a specific taxonomic level.
        \item Chemical metadata in \emph{chemicals.txt} is handled similarly, the data includes chemical name and a (compound) ECOTOX group. 
    \end{enumerate}
    \item The effect data consist of two parts, a test definition and results associated with that test. Note that a test can have multiple results. An example of triples associated with a test is shown in Figure \ref{fig:ecotox_kg_ex}.
    \begin{enumerate}
        \item The important aspects of a test is the compound and the species used, other columns include metadata, but these are optional and often empty. Each result gives an endpoint, an effect (\eg chronic or mortal), and a concentration and unit at which the endpoint and effect where recorded.
        \item We construct a node of type result (\eg \texttt{et:result/001}) and link each result component to it.
    \end{enumerate}
\end{enumerate}

For the units in the effect data, \eg chemical concentrations (mg/L, mol/L, mg/kg, etc.), we reuse the \texttt{QUDT}\footnote{\url{http://qudt.org/1.1/schema/qudt\#}} ontologies. Where a unit is not defined, such as mg/L, we define it as shown in Listing \ref{lst:unit_example}. 

\begin{code}
\begin{minted}{emacs}
@prefix rdf:     <http://www.w3.org/1999/02/22-rdf-syntax-ns#> .
@prefix rdfs:    <http://www.w3.org/2000/01/rdf-schema#> .
@prefix qudt: <http://qudt.org/schema/qudt#> .
@prefix et:   <https://cfpub.epa.gov/ecotox> . 
et:MilligramPerLiter
    rdf:type qudt:MassPerVolumeUnit, qudt:SIDerivedUnit, qudt:DerivedUnit ;
    rdfs:label "Milligram per Liter"^^xsd:string ;
    qudt:abbreviation "mg/L"^^xsd:string ;
    qudt:conversionMultiplier 0.000001 ;
    qudt:conversionOffset 0.0 ;
    qudt:symbol "mg/dm^3"^^xsd:string .
\end{minted}
\vspace{-0.5cm}
\caption{Unit definition of mg/L using \texttt{QUDT}.}
\label{lst:unit_example}
\end{code}

\medskip
\noindent
\textbf{Data alignment API.}\label{sec:alignment}
We use various techniques to align the datasets described above. 

\medskip
\noindent
\textit{ECOTOX-NCBI (Species).} 
There does not exist a complete and public alignment between ECOTOX species and the NCBI taxonomy. 
There exists a partial mapping curated by experts through the ECOTOX search interface,\footnote{\url{https://cfpub.epa.gov/ecotox/search.cfm}} we have gathered a total of 929 mappings for validation purposes.
We use three methods for aligning the two vocabularies.
\begin{enumerate}[\it (i)]
\vspace{-0.2cm}
    \item String matching. We use the Levenshtein distance \cite{1966SPhD...10..707L} between labels of the entities. 
    \item LogMap ontology alignment tool \cite{logmap2011,logma_ecai2012}.
    \item AgreementMakerLight (AML) ontology alignment tool \cite{10.1007/978-3-642-41030-7_38}.
\end{enumerate}
As shown in Table~\ref{tab:agreenment}, the methods achieved high recall over the reference mappings, with Levensthein and LogMap covering all and almost all the reference mappings.
Table~\ref{tab:agreenment} also show the disagreement between methods, which we define as $|S_1 \setminus S_2|$, where $S_1$ and $S_2$ are the sets of computed mappings for methods $1$ and $2$. The disagreement shows that even though AML has the lowest recall, it still suggest a large amount of mappings not discovered by the other methods. This suggests that further analysis of the mappings are required and that a consensus among the different systems will be necessary. Note that, both AML and LogMap exploit the semantics of the input knowledge graphs and implement mapping repair techniques to minimize logical errors in the integration. Such functionality is missing in pure lexical methods like the Levenshtein distance and may lead to noise in the alignment.

\begin{table}[t]
    \centering
    \begin{tabular}{|l|c|c||c|c|c|}\hline
        Method\hspace{5pt} & \hspace{5pt}\#mappings\hspace{5pt} & \hspace{5pt}Recall\hspace{5pt} & \hspace{5pt}Levensthein\hspace{5pt} & \hspace{5pt}LogMap\hspace{5pt} & \hspace{5pt}AML\hspace{5pt} \\\hline 
        Levenshtein\hspace{5pt} & 14915 & 1.0 & 0 & 6711 & 12611\\
        LogMap & 12612 & 0.99 & 4408 & 0 & 10373  \\
        AML & 4299 & 0.71 & 1995 & 2060 & 0\\\hline 
        Intersection & 1979 & 0.71 &  &  &  \\\hline   
    \end{tabular}
    \medskip
    \caption{Alignment results (left of double line) and disagreement between methods (right of double line). Intersection is the mappings which are common across all methods.}\label{tab:agreenment}
\end{table}



\medskip
\noindent
\textit{NCBI-Wikidata (Species).}
The use of more external sources (\eg EOL, MeSH, Freebase) requires a mapping from NCBI identifiers. We construct equivalence triples between NCBI identifiers and Wikidata entities using query shown in Listing \ref{lst:ncbi2wikidata}. This query is then used on the Wikidata endpoint.\footnote{\url{https://query.wikidata.org/sparql}}

\medskip
\noindent
\textit{ECOTOX-Wikidata (Compounds).}
By mapping ECOTOX chemical identifiers (CAS) to Wikidata entities, we enable the use of a vast external chemical datasets, \eg PubChem, ChEBI, KEGG, ChemSpider, MeSH, UMLS, to name a few. The construction of equivalence triples is shown in Listing \ref{lst:cas2inchikey}. 


\begin{code}
    \begin{minted}{sparql}
PREFIX owl: <http://www.w3.org/2002/07/owl#> .
PREFIX wdt: <http://www.wikidata.org/prop/direct/> .
CONSTRUCT {?ncbitaxon owl:sameAs ?taxon .} 
WHERE {
    ?taxon wdt:P685 ?ncbi .
    BIND(IRI(
        CONCAT("https://www.ncbi.nlm.nih.gov/taxonomy/taxon/",
                ?ncbi)) AS ?ncbitaxon)
    }
    \end{minted}
    \vspace{-0.5cm}
    \caption{Construct mapping between NCBI and Wikidata.}
    \label{lst:ncbi2wikidata}
\end{code}
\begin{code}
    \begin{minted}{sparql}
PREFIX owl: <http://www.w3.org/2002/07/owl#> .
PREFIX wdt: <http://www.wikidata.org/prop/direct/> .
CONSTRUCT {?etcompound owl:sameAs ?compound .} 
WHERE {
    ?compound wdt:P231 ?cas .
    BIND(IRI(
        CONCAT("https://cfpub.epa.gov/ecotox/chemical/",
                REPLACE(?cas,'-',''))) AS ?etcompound)
    }
    \end{minted}
    \vspace{-0.5cm}
    \caption{Construct mapping between ECOTOX and Wikidata.}
    \label{lst:cas2inchikey}
\end{code}




\subsection{Accessing TERA} 
\label{sec:dataaccess}

The knowledge in TERA can be accessed via SPARQL queries or via the predefined APIs introduced in Section \ref{sec:datapreparation}. 
The (final) output will depend on the required task, and can be given either as a graph or in tabular format.

\medskip
\noindent
\textbf{SPARQL queries.}
For researchers competent in SPARQL the most powerful method for accessing data in TERA is via SPARQL queries. TERA provides an improved and intuitive method for accessing effect data over the current tabular data base structure. 
We will here give an example of the usability of TERA in extracting data for a risk assessment case study. 


As an example, we have gathered water samples from the inner \emph{Oslofjord}. 
Then, we can extract the compounds and concentrations, at which, the species in the \emph{Oslofjord} experience lethal effects, as shown in Listing \ref{lst:example}. The concentrations can then be compared with water samples (\emph{exposure}) to see if the populations are at risk from contaminants.\footnote{The comparison can be done with another (case study) API. However, this uses only private data and therefore is not included here.}

\begin{code}
    \begin{minted}{sparql}
PREFIX eol: <http://eol.org/schema/terms/> .
PREFIX et:  <https://cfpub.epa.gov/ecotox/> .
SELECT ?s ?c ?conc ?concunit 
WHERE {
    ?s  eol:endemicTo [ rdfs:label "Oslofjorden"@no ] .
    _:b a et:Test ;
        et:species ?s .
        et:compound ?c .
        et:hasResult [
            et:endpoint et:LC50 ;
            et:effectType et:ACUTE ;
            et:concentration [ rdf:value ?conc ;
                               unit:units ?concunit ] .
        ]
} 
    \end{minted}
    \vspace{-0.5cm}
    \caption{Query to select all species, compounds, and concentrations and unit, where the species is endemic to the \emph{Oslofjord}.}
    \label{lst:example}
\end{code}

\medskip
\noindent
\textbf{Predefined APIs.}
In addition to SPARQL queries for extracting data from the knowledge graph, the TERA APIs povide predefined methods which enable access to the data without being proficient in SPARQL,\footnote{Methods are, for the most part, abstractions of SPARQL queries.} but rather prefer a scripting language (here, we use Python).
\begin{enumerate}
    \item In addition to classification, sibling, and name queries, the Species API has methods for fuzzy querying of identifiers based on close matched names. This is a necessary feature, since the name definition may vary from user to user. 
    \item As mentioned before, not all chemical features are included in the TERA knowledge graph, purely for practical reasons. Therefore, fetching features from PubChem is a method in the API. We also include methods for other properties available in PubChem, such as chemical fingerprints, which is a string of bits representing the presence or absence of selected chemical properties. 
    \item For convenience, the Chemical API has methods which wrap SPARQL queries over endpoints or local instances of PubChem, ChEBI or MeSH. This also includes methods for easily converting between identifiers.
    \item The Effect API has also methods that wrap SPARQL queries over the generated knowledge graph. 
\end{enumerate}

\section{TERA Applications}
\label{sec:applications}
In this section, we describe potential uses of TERA that complement the data access application in the ecotoxicological domain. 

\medskip
\noindent
\textbf{Benchmarking embedding models.}
TERA can be used to benchmark existing knowledge graph embedding models for the specific task of effect prediction.
This task uses the three parts (effect, chemical, and taxonomy) separately, where the chemical and taxonomy knowledge graphs are embedded and thereafter used in a prediction model. 
%
In \cite{Myklebust2019KnowledgeGE} we showed that the use of knowledge graph embedding models drastically improved prediction results over a deterministic graph distance approach. However, the work also showed that using embedding models in this novel setting revealed a few shortcomings. First, TERA is majority hierarchical, which could not be properly represented in the embeddings. Efforts have been made to represent hierarchies (\eg Poincare embeddings \cite{nickel2017poincar}), however, we are not aware of models capable of simultaneous capture of horizontal and vertical relations \cite{city23181}. Second, sparsity has a large effect on the performance of knowledge graph embedding models \cite{pujara-etal-2017-sparsity}. Table \ref{tab:stats} show the sparsity of common benchmark datasets and TERA (literals removed). TERA (ECOTOX) and TERA (NCBI) refers to the parts of the knowledge graph that was generated from ECOTOX and NCBI, respectively. TERA (full) contains all data in the materialized snapshot from \cite{tera_kg}.
We follow \cite{pujara-etal-2017-sparsity} and calculate the relational, ${RD}=|T|/|R|$, and entity density, ${ED}=|T|/|E|$, where $T$, $R$, and $E$ are the set of triples, relations, and entities in the knowledge graphs respectably. In addition, we calculate the absolute density of the graph, which is $|T|/(|E|(|E|-1))$. This is the ratio of edges to the maximum number of edges possible in a simple directed graph \cite{doi:10.1137/0720013}.

In Table \ref{tab:stats} we can see that TERA (all) has a high RD, which is down to the mostly hierarchical structure which use the \texttt{rdfs:subClassOf} relation. We see that TERA has similar ED as WN18RR which is a datasets where embedding models has less predictive performance than on \eg FB15k \cite{dettmers2018}. The absolute density of TERA is one order of magnitude lower than WN18 and YAGO3-10.
This makes TERA a very challenging knowledge graph for the next generation of embedding models.

\begin{table}[t]
    \centering
    \begin{tabular}{|l|c|c|c|}
        \hline 
        Dataset & \hspace{5pt}Relational density\hspace{5pt} & \hspace{5pt}Entity density\hspace{5pt} & \hspace{5pt}Absolute density\hspace{5pt} \\\hline 
        TERA (ECOTOX)\hspace{5pt} & $354k$ & $4$ & $1.1\times 10^{-6}$\\
        TERA (NCBI) & $433k$ & $12$ & $2.9\times 10^{-6}$ \\ 
        TERA (full) & $42k$ & $9$ & $1.1\times 10^{-6}$ \\\hline
        YAGO3-10 & $29k$ & $18$ & $7.1\times 10^{-5}$\\
        FB15k & $359$ & $65$ & $2.1\times 10^{-3}$\\
        FB15k-237 & $1148$ & $38$ & $1.3\times 10^{-3}$\\
        WN18 & $7858$ & $7$ & $8.4\times 10^{-5}$\\
        WN18RR & $3$ & $4$ & $5.5\times 10^{-5}$\\\hline
    \end{tabular}
    \medskip
    \caption{Densities of benchmark datasets.}
    \label{tab:stats}
\end{table}

\medskip
\noindent
\textbf{TERA as background knowledge.}
The prediction problem above use the knowledge graph outright. 
However, using TERA as background knowledge where other methods for extrapolating toxicity of chemicals exists is a possible application. 
These methods often use chemical features, images, fingerprints and so on as input, and machine learning methods such as Convolutional Neural Networks and Random Forests as prediction models \cite{Wu2018,yang2018silico}. These models are often uninterpretable, and the predictions lack domain explanations. For machine learning tasks such as preprocessing, feature extraction, transfer and zero/few-shot learning TERA can provide context. Furthermore, the knowledge graph is a possible source for the (semantic) explanation of the predictions (\eg \cite{DBLP:journals/corr/abs-1805-10587}).


\medskip
\noindent
\textbf{Alignment between ECOTOX and NCBI.}
As mentioned in Section \ref{sec:datapreparation}, there does not exist a complete and public alignment between ECOTOX species and the NCBI taxonomy. 
Therefore the computed mappings can also be seen as a very relevant resource to the ecotoxicology community. 
%
The used alignment techniques achieve high scores for recall over the available (incomplete) reference mappings. 
However, aligning such large and challenging datasets requires preprocessing before ontology alignment systems can cope with them.
We removed all nodes which did not share a word (or shared only a stop word) in labels across the two taxonomies. 
This quartered the size of ECOTOX and reduced NCBI 50 fold. However, the possible alignment between entities without labels is lost when reducing the dataset size. 
Thus, the alignment of ECOTOX and NCBI has the potential of becoming a new track of the Ontology Alignment Evaluation Initiative (OAEI) \cite{oaei2018} to push the limits of large scale ontology alignment tools. Furthermore, the output of the different OAEI participants could be merged into a rich consensus alignment that could become the reference to integrate ECOTOX~and~NCBI. 


\section{Discussion and Conclusion}
\label{sec:discussion}
We have created a knowledge graph called TERA and accompanying tools. This knowledge graph aims at covering the knowledge and data relevant to the ecotoxicological domain. 
We have also shown the applications of the knowledge graph, including data retrieval and effect prediction. These applications show the benefits of having a integrated view of the different knowledge and data sources.

\medskip
\noindent 
\textbf{Knowledge graph.}
%
The creation of TERA is of great importance to future effect modelling and computational risk assessment approaches within ecotoxicology,
whose
strategic goal is designing and developing prediction models to assess the hazard and risks of chemicals and 
their mixtures
where traditional laboratory data cannot easily be acquired.
Different knowledge and data sources are integrated into TERA, which aims at consolidating the relevant information to the ecological risk assessment domain. 
The adaption of a RDF-based knowledge graph enables the use of an extensive range of Semantic Web infrastructure (\eg reasoning engines, ontology alignment systems, SPARQL query engines). The accompanying tools enables us to draw conclusions on the effect data from background knowledge, and extrapolate on it.
TERA enables an integrated and semantic access across data sets, and facilitate resource-effective and transparent approaches to optimise this work. Moreover, the contribution is in line with a larger shift in ecological risk assessment towards the use of artificial intelligence~\cite{WITTWEHR2019100114}.

\medskip
\noindent 
\textbf{Value.}
The data integration efforts and the construction of the TERA knowledge graph goes in line with visions in the computational risk assessment communities (\eg Norwegian Institute for Water Research's Computational Toxicology Program (NCTP)), where increasing the availability and accessibility of knowledge enables optimal decision making. For the semantic web community TERA provides a unique dataset which can be used to benchmark new solutions for knowledge graph embedding or in prediction problems.

\medskip
\noindent 
\textbf{Resources.} 
The knowledge graph is available for download from Zenodo \cite{tera_kg}. This download includes all necessary links to the PubChem, ChEMBL, MeSH, EOL, and Wikidata datasets. We also provide a materialized snapshot of TERA.
The APIs are available from GitLab \cite{tera_apis} 
Moreover, given the frequency with which ECOTOX is updated (quarterly), the repository also contains a script for updating the TERA knowledge graph. 

\medskip
\noindent 
\textbf{Maintenance.}
The construction of TERA is a part of a PhD project.
After the finalization of the PhD, there are already plans to maintain and evolve TERA within the Norwegian Institute for Water Research (NIVA) as the use of TERA falls into one of the main research lines of NIVA's Computational Toxicology Program (NCTP). We also expect an engagement from the ecotoxicology community since there is a growing interest in applying artificial intelligence solutions~\cite{WITTWEHR2019100114}.




\section*{Acknowledgements}
This work is supported by grant 
272414 from the Research Council of Norway (RCN), the MixRisk project (RCN 268294), the AIDA project, 
The Alan Turing Institute under the EPSRC
grant EP/N510129/1, 
the SIRIUS Centre for Scalable Data Access (RCN 237889),
the Royal Society, EPSRC projects DBOnto, $\text{MaSI}^{\text{3}}$ and $\text{ED}^{\text{3}}$, and is organized under the Computational Toxicology Program (NCTP) at NIVA.
We would also like to thank Martin Giese and Zofia C. Rudjord for their contribution in different stages of this project.

\bibliographystyle{splncs04}
\bibliography{bibliography}

\begin{thebibliography}{10}

\bibitem{Myklebust2019KnowledgeGE}
E.~B. Myklebust et~al.
\newblock {Knowledge Graph Embedding for Ecotoxicological Effect Prediction}.
\newblock In {\em Int'l Sem. Web Conf. (ISWC)}, 2019.

\bibitem{tera_kg}
E.~B. Myklebust et~al.
\newblock {Toxicological Effect and Risk Assessment (TERA) Knowledge Graph
  (Version 1.0.0) [Data set]}, Dec 2019.
\newblock \url{https://doi.org/10.5281/zenodo.3559865}.

\bibitem{tera_apis}
E.~B. Myklebust et~al.
\newblock {RAPPT: APIs for (pre)processing ecological risk assessment data and
  creating TERA}, Dec 2019.
\newblock \url{https://gitlab.com/Erik-BM/rappt}.

\bibitem{Pradeep2016}
P.~Pradeep et~al.
\newblock {An ensemble model of QSAR tools for regulatory risk assessment}.
\newblock {\em Journal of cheminformatics}, 8:48--48, 2016.

\bibitem{doi:10.1002/qsar.200710099}
T.~I. Netzeva et~al.
\newblock Review of (quantitative) structure–activity relationships for acute
  aquatic toxicity.
\newblock {\em QSAR \& Combinatorial Science}, 27(1):77--90, 2008.

\bibitem{WU201067}
S.~Wu et~al.
\newblock A framework for using structural, reactivity, metabolic and
  physicochemical similarity to evaluate the suitability of analogs for
  sar-based toxicological assessments.
\newblock {\em Regulatory Toxicology and Pharmacology}, 56(1):67 -- 81, 2010.

\bibitem{j.websem510}
H.~Arnaout and S.~Elbassuoni.
\newblock {Effective Searching of RDF Knowledge Graphs}.
\newblock {\em Web Semantics: Science, Services and Agents on the World Wide
  Web}, 48(0), 2018.

\bibitem{jbs2018}
A.~Agibetov et~al.
\newblock Supporting shared hypothesis testing in the biomedical domain.
\newblock {\em J. Biomedical Semantics}, 9(1):9:1--9:22, 2018.

\bibitem{om2013}
J.~Euzenat and P.~Shvaiko.
\newblock {\em Ontology Matching, Second Edition}.
\newblock Springer, 2013.

\bibitem{ecotox}
{U.S. EPA}.
\newblock {ECOTOXicology knowledgebase (ECOTOX)}, 2019.

\bibitem{wikidata2014}
D.~Vrandecic and M.~Kr{\"{o}}tzsch.
\newblock Wikidata: a free collaborative knowledgebase.
\newblock {\em Commun. {ACM}}, 57(10):78--85, 2014.

\bibitem{Willighagen2011}
E.~Willighagen.
\newblock {InChIKey collision: the DIY copy/pastables}, 2011.

\bibitem{Heller2015}
S.~R. Heller et~al.
\newblock {InChI, the IUPAC International Chemical Identifier}.
\newblock {\em Journal of Cheminformatics}, 7(1):23, 2015.

\bibitem{pubchem}
S.~Kim et~al.
\newblock {{PubChem 2019 update: improved access to chemical data}}.
\newblock {\em Nucleic Acids Research}, 47(D1):D1102--D1109, 10 2018.

\bibitem{ncbi}
E.~W. Sayers et~al.
\newblock {{Database resources of the National Center for Biotechnology
  Information}}.
\newblock {\em Nucleic Acids Research}, 37(suppl\_1):D5--D15, 10 2008.

\bibitem{eol}
C.~S. Parr et~al.
\newblock The encyclopedia of life v2: Providing global access to knowledge
  about life on earth., 2014.

\bibitem{1966SPhD...10..707L}
V.~I. {Levenshtein}.
\newblock {Binary Codes Capable of Correcting Deletions, Insertions and
  Reversals}.
\newblock {\em Soviet Physics Doklady}, 10:707, Feb 1966.

\bibitem{logmap2011}
E.~Jim{\'{e}}nez{-}Ruiz and B.~{Cuenca Grau}.
\newblock {LogMap: Logic-Based and Scalable Ontology Matching}.
\newblock In {\em 10th International Semantic Web Conference}, pp. 273--288,
  2011.

\bibitem{logma_ecai2012}
E.~Jim{\'{e}}nez{-}Ruiz et~al.
\newblock Large-scale interactive ontology matching: Algorithms and
  implementation.
\newblock In {\em the 20th European Conference on Artificial Intelligence
  (ECAI)}, pp. 444--449. IOS Press, 2012.

\bibitem{10.1007/978-3-642-41030-7_38}
D.~Faria et~al.
\newblock The agreementmakerlight ontology matching system.
\newblock In R.~Meersman et~al., editors, {\em On the Move to Meaningful
  Internet Systems: OTM 2013 Conferences}, pp. 527--541, Berlin, Heidelberg,
  2013. Springer Berlin Heidelberg.

\bibitem{nickel2017poincar}
M.~Nickel and D.~Kiela.
\newblock {Poincaré Embeddings for Learning Hierarchical Representations},
  2017.

\bibitem{city23181}
O.~M. Holter et~al.
\newblock Embedding owl ontologies with owl2vec.
\newblock {\em CEUR Workshop Proceedings}, 2456:33--36, January 2019.

\bibitem{pujara-etal-2017-sparsity}
J.~Pujara et~al.
\newblock Sparsity and noise: Where knowledge graph embeddings fall short.
\newblock In {\em Proceedings of the 2017 Conference on Empirical Methods in
  Natural Language Processing}, pp. 1751--1756, Copenhagen, Denmark, September
  2017. Association for Computational Linguistics.

\bibitem{doi:10.1137/0720013}
T.~F. Coleman and J.~J. Moré.
\newblock Estimation of sparse jacobian matrices and graph coloring blems.
\newblock {\em SIAM Journal on Numerical Analysis}, 20(1):187--209, 1983.

\bibitem{dettmers2018}
T.~Dettmers et~al.
\newblock Convolutional 2d knowledge graph embeddings.
\newblock 02 2018.

\bibitem{Wu2018}
Y.~Wu and G.~Wang.
\newblock {Machine Learning Based Toxicity Prediction: From Chemical Structural
  Description to Transcriptome Analysis}.
\newblock {\em International journal of molecular sciences}, 19(8):2358, Aug
  2018.

\bibitem{yang2018silico}
H.~Yang et~al.
\newblock In silico prediction of chemical toxicity for drug design using
  machine learning methods and structural alerts.
\newblock {\em Frontiers in chemistry}, 6:30, 2018.

\bibitem{DBLP:journals/corr/abs-1805-10587}
F.~L{\'{e}}cu{\'{e}} and J.~Wu.
\newblock Semantic explanations of predictions.
\newblock {\em CoRR}, abs/1805.10587, 2018.

\bibitem{oaei2018}
A.~Algergawy et~al.
\newblock Results of the ontology alignment evaluation initiative 2018.
\newblock In {\em 13th International Workshop on Ontology Matching}, pp.
  76--116, 2018.

\bibitem{WITTWEHR2019100114}
C.~Wittwehr et~al.
\newblock Artificial intelligence for chemical risk assessment.
\newblock {\em Computational Toxicology}, pp. 100114, 2019.

\end{thebibliography}

\end{document}